\documentclass[5p,times]{elsarticle}

\usepackage{amssymb}

\usepackage{rotating}
\usepackage{multirow}
\usepackage{amsmath}
\usepackage{graphicx}
\usepackage{xcolor}
\usepackage{multirow}
\usepackage{algorithm} 
\usepackage{algpseudocode} 

\journal{Future Generation Computer Systems}

\begin{document}

\begin{frontmatter}



\title{Grassroots Operator Search for Model Edge Adaptation}


\author[label1]{Hadjer Benmeziane}
\author[label2]{Kaoutar El Maghraoui}
\author[label1]{Hamza Ouarnoughi}
\author[label1]{Smail Niar}
 
 \affiliation[label1]{organization={Univ. Polytechnique Hauts-de-France, CNRS UMR8201 LAMIH},
             city={Valenciennes},
             postcode={59300},
             country={France}}

 \affiliation[label2]{organization={IBM T. J. Watson Research Center, Yorktown Heights},
             city={New York},
             postcode={10598},
             country={USA}}
             
\begin{abstract}
Hardware-aware Neural Architecture Search (HW-NAS) is increasingly being used to design efficient deep learning architectures. An efficient and flexible search space is crucial to the success of HW-NAS. Current approaches focus on designing a macro-architecture and searching for the architecture's hyperparameters based on a set of possible values. This approach is biased by the expertise of deep learning (DL) engineers and standard modeling approaches. 
In this paper, we present a Grassroots Operator Search (GOS) methodology. Our HW-NAS adapts a given model for edge devices by searching for efficient operator replacement. We express each operator as a set of mathematical instructions that capture its behavior. The mathematical instructions are then used as the basis for searching and selecting efficient replacement operators that maintain the accuracy of the original model while reducing computational complexity. Our approach is grassroots since it relies on the mathematical foundations to construct new and efficient operators for DL architectures.  
We demonstrate on various DL models, that our method consistently outperforms the original models on two edge devices, namely Redmi Note 7S and Raspberry Pi3, with a minimum of 2.2x speedup while maintaining high accuracy. Additionally, we showcase a use case of our GOS approach in pulse rate estimation on wristband devices, where we achieve state-of-the-art performance, while maintaining reduced computational complexity, demonstrating the effectiveness of our approach in practical applications.
\end{abstract}




\end{frontmatter}


\section{Introduction}
Hardware-aware Neural Architecture Search (HW-NAS) is a technique to design efficient Deep Learning (DL) architectures for different tasks such as image classification~\cite{fbnetv3} and object detection~\cite{nasfos} in computer vision.

HW-NAS follows three steps. First, a \textit{search space} is defined with a set of possible DL architectures. 
Second, a multi-objective \textit{search strategy} is implemented to explore the search space to find the best architecture. 
The search strategy uses an \textit{evaluation methodology} to evaluate each sampled architecture against different objectives such as accuracy, latency, and energy consumption.
Finally, the architecture that presents the best objectives' trade-off is defined as the \textit{"best"} architecture. 
In this paper, the term \textit{architecture} refers to the DL architecture and the term \textit{architecture performance} refers to combining task-performance metrics, such as the accuracy or average precision, and the hardware efficiency computed using latency, energy consumption, and memory occupation of a sampled architecture.  

The definition of the search space is a critical step in NAS. It determines the range of possible architectures and can significantly impact the final performance. The size of the search space matters. A large search space hinders the exploration but diversifies the results. In contrast, a small search space restricts architectural diversity. 
Currently, there are three primary approaches to define the search space in HW-NAS~\cite{survey}:  
\begin{enumerate}
    \item 
    \textit{Cell-based search space}, which involves searching for a repeated cell, also called \textit{block}, within a pre-defined macro-architecture. 
    The cell is defined by a list of operators, such as convolution and batch normalization and an adjacency matrix that defines the connections between the operators. NAS-Bench-101~\cite{101} is a common NAS benchmark designed using a cell-based search space.
    \item
    \textit{Hierarchical search space}~\cite{hierarchical} extends the cell-based approach by selecting the operators composing the cell, defining the cell-level connections, and merging multiple cells. 
    \item
    \textit{Supernetwork search space}~\cite{darts}, in which each architecture is represented as a subgraph within a larger and more complex network called the \textit{supernetwork}. The weights of the supernetwork are shared among all subgraphs, allowing the subnetworks to share computation and enabling efficient exploration of the search space. The supernetwork is called an over-parameterized network. The subgraphs can differ in terms of their connectivity, layer types, layer sizes, and other architectural hyperparameters.
\end{enumerate}

A prevalent limitation of such definitions is the bias introduced by the dependence on human-designed architectures, which restricts the search algorithms from exploring novel and innovative operations and architectures. This bias towards previously handcrafted architectures hinders the discovery of more efficient and effective models for specific tasks. Consequently, there is a need to develop novel methodologies that can help discover more optimized architectures and operations that can perform well on various devices and scenarios without relying on pre-existing models. Such methodologies would be the holy grail of NAS, as they would enable the creation of truly novel architectures that can push the limits of deep learning performance even further.

One solution defines a giant search space where the architecture and operations are generated from scratch and then evaluated based on their performance. However, given the vast search space, such an approach requires a massive amount of computational resources and is often infeasible for practical use. AutoML-Zero~\cite{automlzero}, for example, presents a strategy capable of defining the architecture and the training procedure from standard mathematical operations using reinforcement learning. 
This approach breaks the innovation barrier for NAS but at a significant time complexity price.  
Due to the highly complex search, AutoML-Zero only achieves linear regression on the MNIST dataset, which is impractical for complex and real-world datasets.

Selecting the right set of operators for a specific task is crucial, however, the actual implementation of the operator can also greatly impact the hardware efficiency of the DL model. To overcome this challenge, recent works have focused on using DL compilers~\cite{tvm, tiramisu} that can automatically select the most efficient implementation and optimization for a given hardware. These compilers use techniques such as code generation and optimization, which automatically translate the high-level DL operators to hardware-specific low-level code to improve the efficiency of DL models on different hardware devices such as edge devices. The use of deep learning compilers highlights the importance of not only selecting the right operator but also optimizing its implementation to achieve the best possible hardware performance. MCUNet~\cite{mcunet} combines the use of NAS and \textit{TinyEngine}, a deep learning compiler for microcontrollers, to efficiently look for the best architecture and its hardware efficient implementation in an iterative manner. However, their search space is limited to a set of standard DL operators, whose implementations are not optimized for edge or resource-constrained devices.

This paper presents a search algorithm that adapts the architecture to edge devices without previous human experience. To overcome the time complexity of AutoML-Zero, we apply our search algorithm on a specific layer at each iteration. In the first step, we analyze each layer's latency and memory occupancy distributions in a given model. We consider a model as a set of layers such as convolution. Each layer corresponds to a sequence of operators implemented by a graph of mathematical instructions. Table~\ref{tab:math_ops} gives the list of mathematical instructions considered in this work. In the second step, the most inefficient layer is optimized. Costly operators in this layer are replaced by efficient operators. An operator is a set of mathematical instructions that capture its behavior. For example, standardization is defined by subtracting the mean of the input over a mini-batch and dividing it by the standard deviation of that input.
The mathematical instructions are then used as a basis for searching and selecting efficient replacement operators that maintain the accuracy of the original model while reducing computational complexity. 

We repeat these two steps until we find an architecture suited for the targeted edge device without dropping accuracy. 
Our technique aims to break the time-consuming barrier of non-restrictive search spaces while searching for new and innovative architectural designs. 

We summarize the contributions described in this paper as follows.
\begin{itemize}
    \item We present a new adaptation methodology via operator replacement. We replace the most hardware-inefficient layer iteratively by building new operators from scratch with minimal human bias. 
    \item We develop an optimized multi-objective evolutionary search algorithm that effectively selects the appropriate operators for deploying an efficient architecture on the targeted device. This enables the deployment of deep learning models on edge devices with improved efficiency and without sacrificing accuracy. 
\end{itemize}

Our methodology has been validated with different types of architectures: Convolutional neural networks (ConvNets) and Vision transformers (ViT). In particular, we identified a novel convolution implementation suitable for Raspberry Pi, is a significant contribution to the field of edge computing. Additionally, we applied our methodology for Pulse Rate estimation with Photoplethysmography (PPG) sensors and achieved state-of-the-art results. Overall, our approach consistently improves the model's hardware efficiency with an average of 2x speedup without any loss in the model's accuracy. These results demonstrate the effectiveness and versatility of our methodology for optimizing DL models for different hardware platforms and applications.

\section{Related Works \& Background}
In this section, we review related works on two different aspects of our methodology: defining fine-grained search spaces and using deep learning compilers for optimizing DL operators in HW-NAS.

\subsection{Fine-grained Search Space for NAS}
The term "\textit{fine-grained search spaces}" refers to search spaces that consist of a set of mathematical and low-level functions. Such search spaces for NAS. The reason is due to the large number of possible operators created from this search space, hindering the exploration. 

AutoML Zero~\cite{automlzero} is the only AutoML tool that defines a search space from basic operators. Their goal is to search for the end-to-end learning pipeline, i.e., from architecture building blocks to optimizing the loss function. This work is a seminal step towards the holy grail of AutoML: automatically designing a network and training pipeline for any given dataset. However, their methodology took a tremendous amount of time to come up with an already human-designed logistic regression. Recently, BANAT~\cite{banat} proposes an algebraic representation of the architecture to enable a more general search space definition. This is a promising direction for efficiently and effectively searching over our huge search spaces. 

Other works~\cite{banat,evonorm, evoprompting} consider modifying a single operator, namely batch normalization. EvoNorms~\cite{evonorm} evolves the normalization operator from basic mathematical functions. They discovered novel implementations and functions for the normalization and activation fusion which improved the overall average precision of multiple standard models. 

Due to their recent application and high time complexity, low-level search spaces are only considered in NAS with a task-specific objective. In other terms, our work is the first to search for adapting the model for resource-constrained devices using a low-level search space. 

\subsection{DL Compilers \& Hardware-aware Neural Architecture Search}
In addition to search algorithms that operate on high-level architectures and operations, several works have explored the optimization of DL models at the hardware and software levels. One approach to this problem is through the use of DL compilers~\cite{tvm, tiramisu}, which automatically optimize the code of a given model for a target hardware platform. These compilers employ a range of techniques, such as graph rewriting, operator fusion, and kernel selection, to reduce memory usage, improve compute performance, and exploit hardware-specific features.
For instance, MCUNet~\cite{mcunet} developed a dedicated compiler that enhances the convolutions with loop tiling. The search iterates over searching for the architecture, then searching for the best optimization and extracting the hardware performance. While this methodology is proven efficient, their search space consists of standard operations which hinder the innovation and adaptation of multiple hardware platforms in HW-NAS.
Other DL compilers such as TVM~\cite{tvm} and Tiramisu~\cite{tiramisu} predict the adequate optimization to apply for a given operator. 

Our primary goal is to tailor the DL architecture to a specific hardware platform by systematically replacing the least-efficient operator in an iterative manner, employing a fine-grained search space. This approach streamlines the resource-intensive process of exploring an extensive search space by concentrating on the adaptation of individual operators step-by-step. 

\subsection{Pulse Rate Estimation}
Pulse rate estimation~\cite{pankaj2022review} has been the subject of extensive research in the field of physiological monitoring. Various approaches utilize photoplethysmography (PPG) signals captured from wearable devices, such as wrist-worn sensors or fingertip sensors, to estimate the pulse rate. Among the state-of-the-art pulse rate estimation models, we compare our results against DeepHeart~\cite{deepheart}, CNN-LSTM~\cite{CNN-LSTM}, and NAS-PPG~\cite{ppgnas}.

DeepHeart uses an ensemble of denoising convolutional neural networks (DCNNs) to denoise contaminated PPG signals that are then passed through spectrum-analysis-based calibration to estimate the final pulse rate. 

CNN-LSTM uses a hybrid convolutional and LSTM neural network. The proposed model is comprised of two convolutional layers, two LSTM layers, one concatenation layer, and three fully connected layers including a softmax.  

NAS-PPG is the first NAS applied to pulse rate estimation. Their search space is defined with a convolutional macro-architecture comprising time-distributed convolutions and two final LSTM layers. Thanks to their automatic search, they provide the best performance on Troika dataset~\cite{troika}. 

This task serves as an excellent validation use case for our methodology, specifically tailored to edge devices, where limited computational resources and power constraints present unique challenges. By effectively optimizing pulse rate estimation models for edge devices, we showcase the practicality and robustness of our approach in overcoming these constraints and meeting the specific requirements of edge environments. 

\section{GOS: Grassroots Operator Search}
Figure~\ref{fig:overview} shows the overall structure of our methodology. 
Given a model, denoted as $m$, our goal is to adapt it to a targeted edge platform. 
We define an \textit{operator} as a collection of operations applied within a layer. These operations can encompass a single layer, as commonly found in deep learning frameworks (e.g., convolution), or a fused layer, such as the combination of ReLU and Batch Normalization (ReLU-BN)~\cite{fusion}. This distinction allows us to work with both individual and composite layer types in our adaptation process.

The process goes through two stages: 
\begin{figure*}
    \centering
    \includegraphics[width=0.8\textwidth]{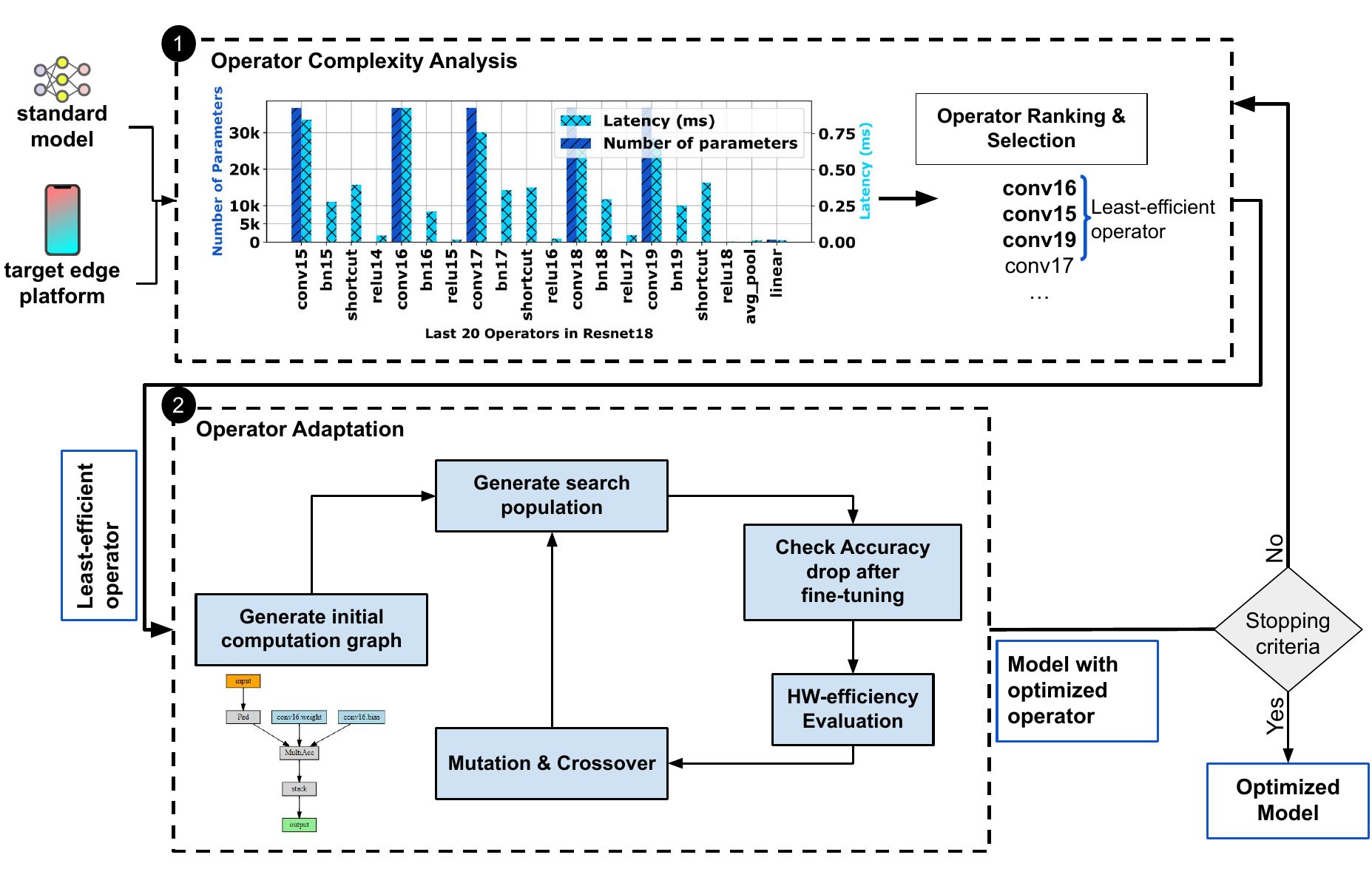}
    \caption{Overview of the Grassroots Operator Search (GOS) framework.}
    \label{fig:overview}
\end{figure*}
\begin{enumerate}
    \item \textit{Operator Complexity Analysis:} First, our process identifies the least efficient operator within the given model ($m$) by conducting $N_i$ inference runs on the target edge device. The efficiency metric is computed with different objectives, such as latency and the number of parameters. The number of parameters reflects the size of the operator. Additional criteria such as energy consumption may be added. Among the list of operators in $m$, the least efficient operator is selected based on algorithm~\ref{alg:selection}. If the model is not deployable on the target platform, i.e., the size of the network exceeds the memory capacity, we select the operator with the highest number of parameters denoted as $num\_param$ in algorithm~\ref{alg:selection}. Otherwise, we rank the architectures with latency and number of parameters in descending order and select the first operator. Our strategy of ranking is as follows: if the architecture is deployable on the target device, the number of parameters is a less important objective, we rank the operators based on the latency and if two operators are of close latencies then we consider the number of parameters. This behavior is checked at each iteration. If more criteria are considered, then the ranking should be multi-objective~\cite{multiobjective}. This operator corresponds to the slowest operator that has the highest number of parameters possible. If an operator is selected, it cannot be selected for another optimization iteration. In ConvNets, it is common knowledge that the least-efficient operator is the convolution. However, according to its input and output shape, the convolution may be optimized differently. To efficiently select the operator to be replaced, we define $N_o$ as the maximum number of similar operators and select the top operators each time. For example, if the $N_o$ least efficient operators are all convolutions, we will replace them all with the same generated optimized operator.  By selecting the top $N_o$ least efficient operators during each iteration, we strike a balance that ensures both effective optimization and a manageable adaptation time complexity.

    \item \textit{Operator Adaptation:} Then, we adapt the selected operator by searching for a variation that can keep the same input and output shapes but optimizes the computations. This phase is done with an evolutionary search on a set of mathematical operations. Section~\ref{sec:search_space} and section~\ref{sec:search_algo} describe the search space and methodology respectively. During the search, only the parameters of the adapted operator are fine-tuned.
\end{enumerate}
The two steps are repeated until satisfactory hardware efficiency is reached or a maximum number of layers have been replaced. 

\begin{algorithm}
    \caption{Least-efficient Operator Selection}
    \label{alg:selection}
    \begin{algorithmic}
        \State \textbf{Input:} Model $m$, Number of inference $N_i$
        \State $is\_deployable \gets deploy(m)$
        \If{not is\_deployable }
            \For{each $o$ in $m$} {\color{blue}{ \# for each operator o get its number of parameters}}
                \State $num\_param[o] \gets number\_of\_params(o)$ \color{black}
            \EndFor
            \State \textbf{return} argmax(num\_param, $N_o$) 
            \State {\color{blue}{\#  return the operator with highest value in num\_param and its number of occurrences $N_o$}}
        \EndIf
        \For{each $o$ in $m$} 
            \State $latency[o] \gets average\_latency(o, N_i)$  {\color{blue}{\#  compute the mean latency of each operator o for $N_i$ inferences}}
            \State $num\_param[o] \gets number\_of\_params(o)$
            
        \EndFor
        \State \textbf{return} Top $N_o$ similar operators 
        \State {\color{blue}{\#  return the operator with highest value in num\_param and its number of occurrences $N_o$}}
    \end{algorithmic}
\end{algorithm} 

\subsection{Operator Search Space}
\label{sec:search_space}
Unlike previous HW-NAS search spaces that are based on pre-defined operator sets, our search space is defined with a set of mathematical operations. The operator is represented with a computation graph. The computation graph is a directed acyclic graph (DAG) with $N$ nodes and $E$ edges. The edges describe the inputs and outputs of each node. Figure~\ref{fig:overview} (step 2) shows an example of such a graph.

Each node in the context can be classified into one of the following three types:
\begin{itemize}
    \item Instruction: This node corresponds to any mathematical instruction in table~\ref{tab:math_ops}. 
    \item Input: This node corresponds to the input feature maps or weights that are given as operands to the instruction node.
    \item Constant: This introduces hyperparameters fixed in the mathematical instruction equation. These constants can be tuned and mutated during the search. 
\end{itemize}

We constrain the generated computation graphs with $1<N<=20$ and $1<E<=25$. These values have been fixed by analyzing standard models' operators. During the generation, the input node is fixed, and its shape is defined by the output of the previous operation in $m$. The output node's shape is also known as it is constrained by the input shape of the next operator in $m$. To ensure a valid network, we optionally add a reshape operation at the end of the computation graph to keep the same output shape as expected by the next operator in the given model. Nodes that can't be reached from the input or that do not have a path to the output are considered unused and therefore pruned from the computation graph. 

Table~\ref{tab:math_ops} shows the basic operations in the search space, including arithmetic, linear algebra, probability, and aggregation operations. The aggregation operations enable to merge between the output of multiple nodes. We include code optimizations such as loop tiling and unrolling as special aggregation functions that are called to optimize the generated operator's code. Note that this is a general application of these optimizations that can be hardware-specifically defined by a compiler. 

{{Note that each operator has a list of hyperparameters dedicated to it. These hyperparameters are illustrated in the equations as constants in table~\ref{tab:math_ops}.}}

\begin{table*}[h!]
    \centering
    \caption{List of mathematical instructions defining the search space}
    \begin{tabular}{c|l|c}
        \hline \hline
        Category & Instruction & Equation \\
        \hline\hline
        Linear Algebra & Matrix multiplication & $\mathbf{C} = \mathbf{A} \mathbf{B}$ \\
         & Matrix addition and subtraction & $\mathbf{C} = \mathbf{A} + \mathbf{B}$ or $\mathbf{C} = \mathbf{A} - \mathbf{B}$ \\
         & Vector multiplication & $\mathbf{c} = \mathbf{A} \mathbf{b}$ \\
         & Matrix inversion & $\mathbf{A}^{-1}$ \\
         & Dot product & $\mathbf{a}^\top \mathbf{b}$ \\
          & Determinant & $\mathrm{det}(\mathbf{A})$ \\
         & Trace & $\mathrm{tr}(\mathbf{A})$ \\
         & Eigenvalues and eigenvectors & $\mathbf{A} \mathbf{v} = \lambda \mathbf{v}$ \\
         & Singular value decomposition (SVD) & $\mathbf{A} = \mathbf{U} \boldsymbol{\Sigma} \mathbf{V}^\top$ \\
         & QR decomposition & $\mathbf{A} = \mathbf{Q} \mathbf{R}$ \\
         & Cholesky decomposition & $\mathbf{A} = \mathbf{L} \mathbf{L}^\top$ \\
         & Matrix pseudoinverse & $\mathbf{A}^\dagger$ \\
         & Matrix rank & $\mathrm{rank}(\mathbf{A})$ \\
         & Hadamard product & $\mathbf{C} = \mathbf{A} \odot \mathbf{B}$ \\
         & Kronecker product & $\mathbf{C} = \mathbf{A} \otimes \mathbf{B}$ \\
         & Outer product & $\mathbf{C} = \mathbf{a} \mathbf{b}^\top$ \\
         & Vector norm & $\|\mathbf{x}\|$ \\
         & Matrix norm & $\|\mathbf{A}\|$ \\
         & Frobenius norm & $\|\mathbf{A}\|_F$ \\
         & Identity matrix & $\mathbf{I}$ \\
         & Zero matrix & $\mathbf{0}$ \\
        \hline \hline
        Calculus & Gradients & $\nabla_\theta L(\theta)$ \\
         & Partial derivatives & $\frac{\partial f}{\partial x}$ \\
         & Chain rule & $\frac{\partial f}{\partial x} = \frac{\partial f}{\partial g} \frac{\partial g}{\partial x}$ \\
        \hline\hline
        Activation Functions & Sigmoid & $\sigma(x) = \frac{1}{1 + e^{-x}}$ \\
         & ReLU & $\mathrm{ReLU}(x) = \max(0, x)$ \\
         & Tanh & $\mathrm{tanh}(x) = \frac{e^x - e^{-x}}{e^x + e^{-x}}$ \\
         & Softmax & $\mathrm{softmax}(x_i) = \frac{e^{x_i}}{\sum_{j=1}^k e^{x_j}}$ \\
        \hline\hline
        Convolution & {{cross-correlation}} & $(f \ast g)(x,y) = \sum_{i=-k}^k \sum_{j=-k}^k f(x-i,y-j) g(i,j)$ \\
        \hline\hline
        Pooling & Max pooling & $\mathrm{maxpool}(x_{i:i+s, j:j+s}) = \max_{m=1}^s \max_{n=1}^s x_{i+m,j+n}$ \\
         & Average pooling & $\mathrm{avgpool}(x_{i:i+s, j:j+s}) = \frac{1}{s^2} \sum_{m=1}^s \sum_{n=1}^s x_{i+m,j+n}$ \\
        \hline\hline
        Probability and Statistics & Probability distributions & $p(x)$ \\
         & Bayesian inference & $p(\theta|x) = \frac{p(x|\theta)p(\theta)}{p(x)}$ \\
        \hline\hline
        Aggregation Function &Summation & $\sum_{i=1}^n x_i$ \\
        &Mean & $\frac{1}{n}\sum_{i=1}^n x_i$ \\
      
        &Maximum & $\max(x_1, x_2, ..., x_n)$ \\
                &Minimum & $\min(x_1, x_2, ..., x_n)$ \\
   
        &Square Root & $\sqrt{x}$ \\
    
        &Concatenation & $\begin{bmatrix} A \ B \end{bmatrix}$\\
        & Weighted Mean & $\frac{\sum_{i=1}^n w_i x_i}{\sum_{i=1}^n w_i}$\\
        \hline\hline
    \end{tabular}
    \label{tab:math_ops}
\end{table*}

\begin{figure}[h]
    \centering
    \includegraphics[width=0.45\textwidth]{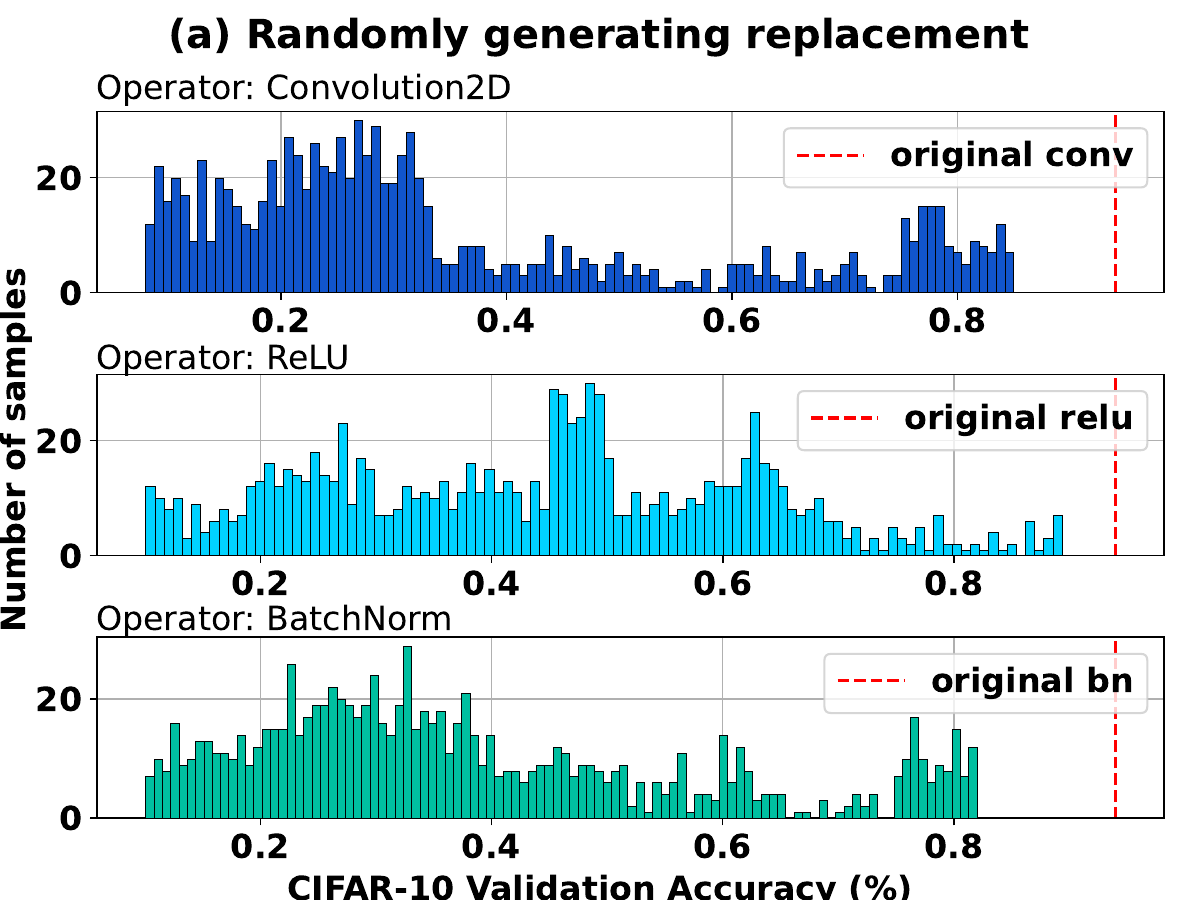}\vspace{1cm}
    \includegraphics[width=0.45\textwidth]{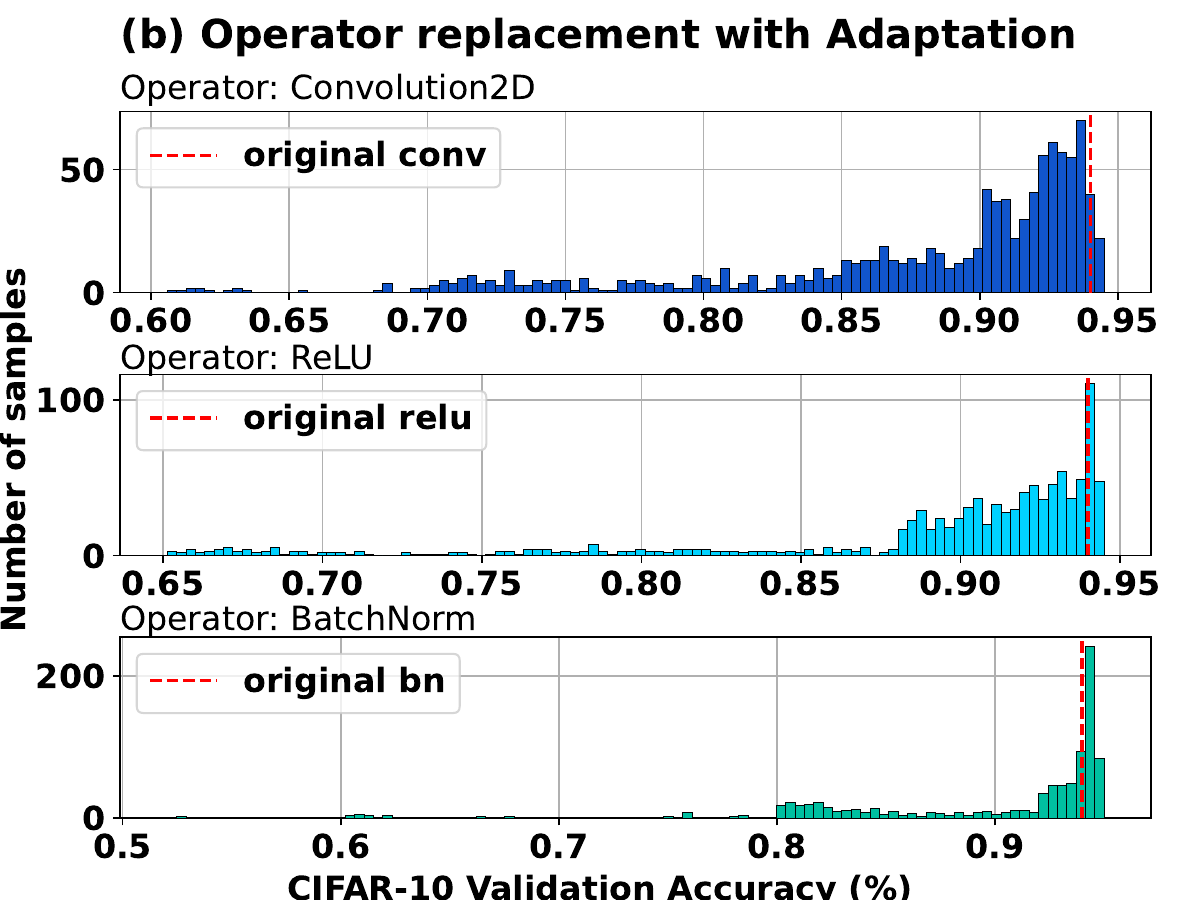}
    \caption{CIFAR-10 accuracy histograms of 1k architectures randomly generated (a) and adapted from the original operator (b). }
    \label{fig:generation}
\end{figure}

{
{
\paragraph{Example of Operator Computation Graph}
In this paragraph, we explain how the convolution 2D is turned into a computation graph. In its simplest form, the convolution 2D can be formulated as in equation~\ref{eq:conv}, where $N$ is the batch size, C denotes the number of channels, H is the height of input planes in pixels, and W is the width in pixels. $in$ and $out$ refer to the input and output respectively. $\ast$ in the equation denotes the cross-correlation operation~\cite{DBLP:conf/cvpr/ValmadreBHVT17}. 

\begin{equation}
    \label{eq:conv}
    conv2D(N, C_{out}) = bias(C_{out}) + \sum_{k=0}^{k = C_{in}-1} weight(C_{out}, k) \ast input(N, k)
\end{equation}

The convolution first splits the input into weight-shaped chunks. We compute the multiply-accumulate of each of these chunks with the weights (i.e., kernels), using the cross-correlation instruction. We then sum up all the multiplied values over the input channels $C_{in}$. Finally, we add the bias to each output channel $C_{out}$. 

To create the computation graph, we divide the equation into instructions found in table~\ref{tab:math_ops}. Figure~\ref{fig:conv2D} shows the complete convolution 2D graph with a 2-dimensional input and 2 kernels. To have a compact and simple graph, we include the constant nodes inside the instruction node as a list of hyperparameters. In the rest of the paper and for the sake of clarity, we use high-level operator names such as Linear for the matrix multiplication between weight and input matrices. 

\begin{figure*}
    \centering
    \includegraphics[width=0.8\textwidth]{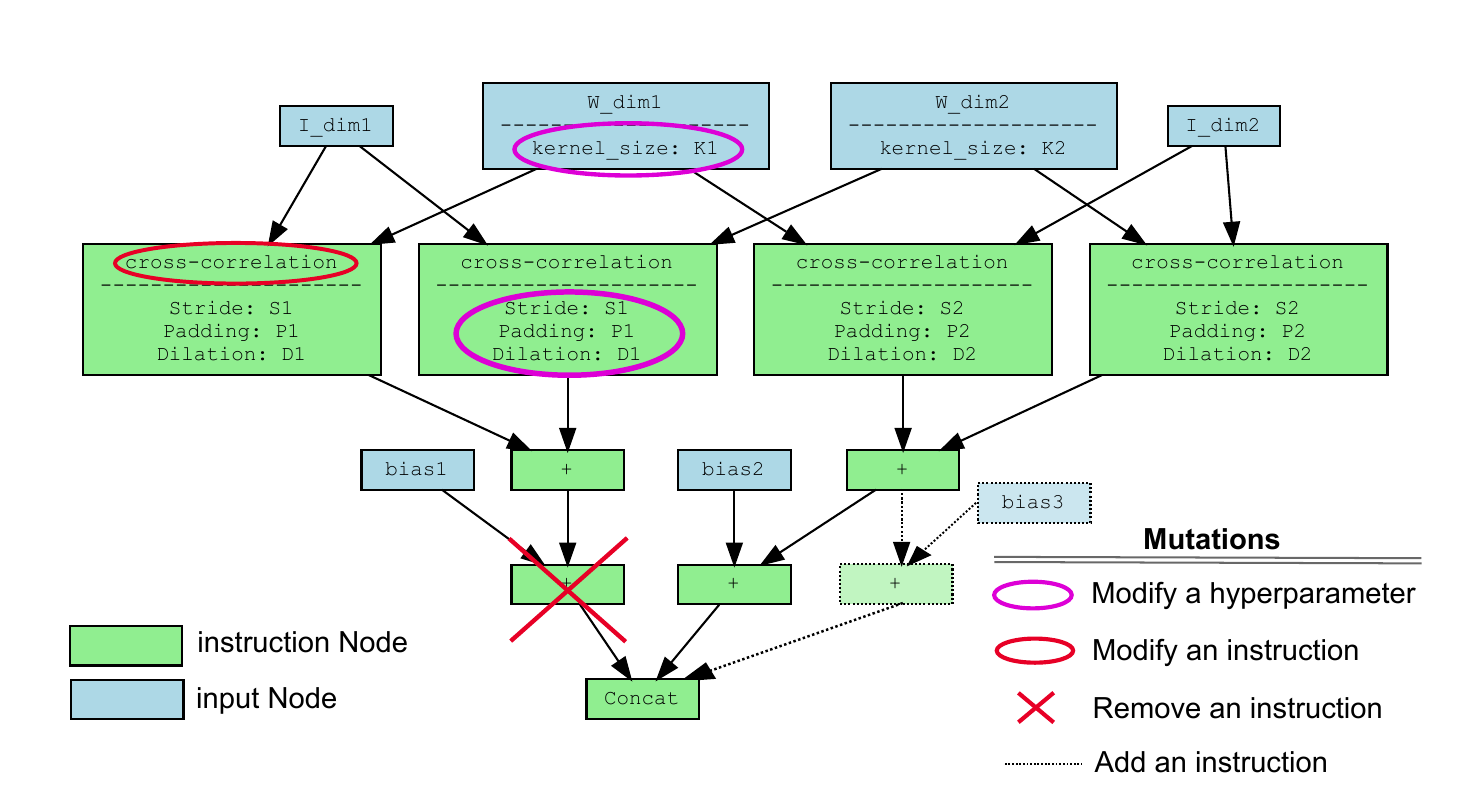}
    \caption{Detailed computation graph of the standard convolution 2D including the possible mutations applied to it.}
    \label{fig:conv2D}
\end{figure*}
}
}

In this search space, we perform small-scale experiments with random sampling to understand its behaviors. The purpose is to measure the sparsity of the search space and to determine the number of valid and accurate operations generated during the exploration. In this experiment, we replace all similar operators at once. For example, we replace all convolutions in the model with a generated replacement. Figure~\ref{fig:generation} (a) shows the results of 1000 randomly generated operator replacements for three operators: Conv2D, max-pooling, and batch normalization, in resnet-18~\cite{resnet}. Random generation, inspired by EvoNorm~\cite{evonorm}, starts from the input node and sequentially selects an operation from the search space. In all the cases, the ImageNet accuracy drops significantly for most of the replacements, which reflects the high sparsity of our search space. In figure~\ref{fig:generation} (b), rather than randomly generating the operator replacement, we start with the original operations but adapt one operation in the computation graph. The adaptation is performed while being aware to keep the same arity and type of arguments for each operation. With adaptation, the results are much closer to the original accuracy of the model but the complexity is modified. 

\subsection{Search Algorithm}\label{sec:search_algo}
Given an operator computation graph, the search algorithm aims at finding a variant that preserves the accuracy of the model with reducing complexity. We rely on an evolutionary algorithm for this purpose. The evolutionary algorithm allows us to handle the sparse search space by exploring a population of valid computation graphs. The computation graph is considered valid if it maintains the shapes of the input and output data and if there exists a path from every intermediate node, including the input node, to the output node. Besides, mutation and crossover provide an efficient way to generate complex adaptations. We use tournament selection which ensures that the best individuals have a higher chance of being selected, while still allowing for some diversity in the population. This helps to prevent premature convergence and promotes the discovery of novel solutions in our large search space. 

\paragraph{Mutations}
The mutation operations involve modifying the computation graph. {{Figure~\ref{fig:conv2D} summarizes the possible mutations applied on the conv2D computation graph. }} Each instruction node in the computation graph is typed with the corresponding type in table~\ref{tab:math_ops}. The most important mutation is modifying any intermediate node with a possible operation. For each operation, we associate a list of possible replacements. The replacement satisfies two constraints: (1) having the same argument's type and arity, (2) the output shape is equal or can be converted to the original output shape by adding a reshape operation. The replacement operation from the list is selected uniformly at random. We also allow for a modification of the aggregation function, and an addition or deletion of a node. When adding or removing a node, we make sure that a path from the input to the output is still possible and that no unused node appears in the graph. 

The mutations also include modifying the hyperparameter of the operator. The hyperparameters are properties associated with a vertex in the computation graph. For each instruction, a list of possible hyperparameters; i.e., constants, is available. For each hyperparameter, we constrain the ranges with specified values obtained from the literature. For example, the output channel size of a convolution may change. This mutation may reduce the accuracy of the model. If this is the case, the operator is invalidated and is not considered in the novel population.

\begin{figure*}
    \centering
    \includegraphics[width=\textwidth]{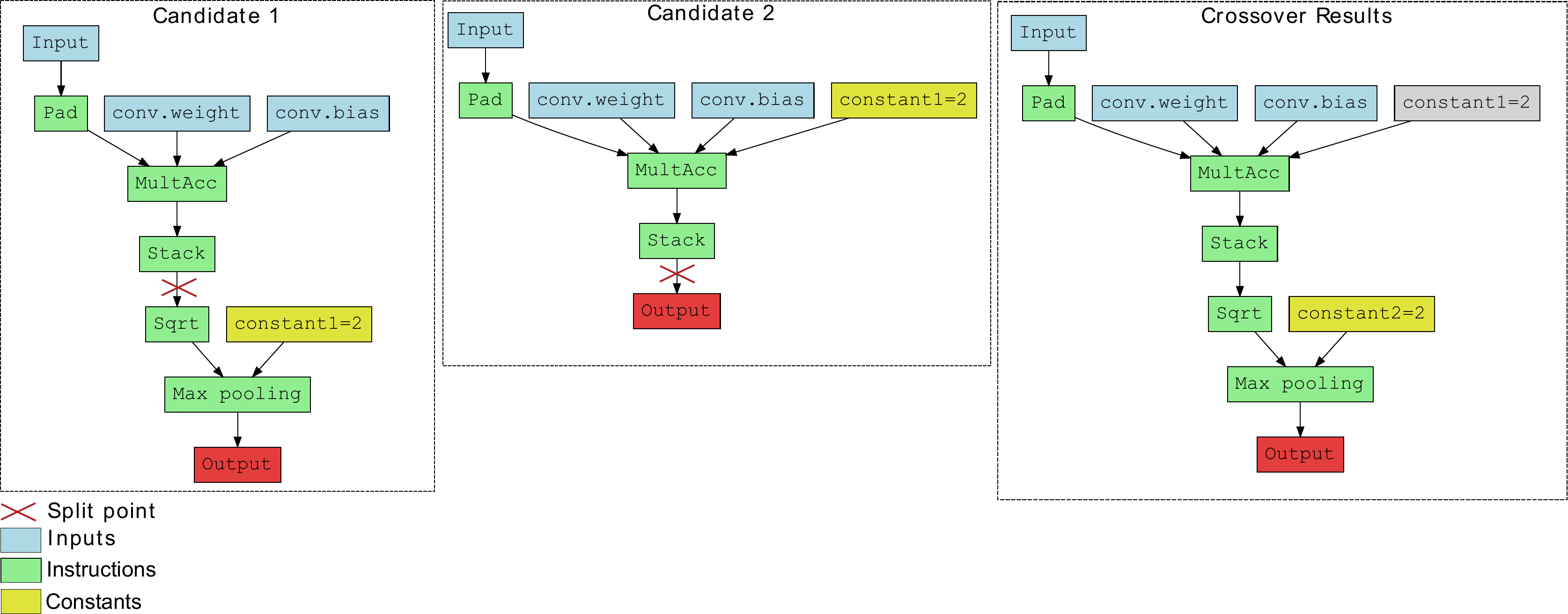}
    \caption{Illustration of the cross-over operation.}
    \label{fig:cross}
\end{figure*}

\paragraph{Crossover}
In general, the crossover is not applied to NAS algorithms. When we consider high-level operators, it is rarely the case to find a splitting point where the shapes fit. However, in our case, the crossover is beneficial and allows more flexibility. Algorithm~\ref{alg:crossover} and figure~\ref{fig:cross} detail the crossover procedure. We perform a crossover between two computation graphs in our population. Because all the variants start from the same point, we have more chances to find a split point. We perform a pre-order traversal of the two computation graphs and store all the possible split points. We randomly select a split point between each pair of computation graphs and generate offspring. 

\begin{algorithm}
    \caption{Crossover procedure 
    }
    \label{alg:crossover}
    \begin{algorithmic}
        \State \textbf{Input:} Two computation graphs of two operators ($o1$ and $o2$)
        \State split\_points = []
        \State Stack s = Stack()
        \State Push ($o1$, $o2$) to s
        {\color{blue}{ \# pre-order traversal of both computation graphs}}
        \While{s not empty}
            \State Pop a node pair ($o1$, $o2$) from the top of the stack
            \While{ $o1$ not empty and $o2$ not empty}
                \State Pop a node pair ($o1$, $o2$) from the top of the stack
            \EndWhile
            \If{shape($o1$.output) == shape($o2$.input)}
                \State Add ($o1$, $o2$) to split\_points {\color{blue}{ \# add to possible split point}}
            \EndIf
            \For{child of $o1$}
                \State Add (child, $o2$) to stack
            \EndFor
            \For{child of $o2$}
                \State Add ($o1$, child) to stack
            \EndFor

        \State Uniformly select ($o1$, $o2$) from split\_points {\color{blue}{ \# randomly select a split points between all the possibilities}}
        \State Perform a merge illustrated in Figure~\ref{fig:cross}
    \EndWhile
    \end{algorithmic}
\end{algorithm} 

\paragraph{Multi-objective Fitness Function}
The evaluation is specific to the given model and task. We do not generalize the resulting operation to multiple standard models because our goal is to adapt the network for a given hardware platform in a practical time. This allows a more flexible and multi-objective fitness function. 

The fitness function evaluates the performance of the adapted operator, formulated in equation~\ref{eq:fitness}. In our methodology, we consider hardware efficiency with multiple objectives. Our definition considers latency and the number of parameters. But one can add other objectives such as energy consumption or memory occupancy. We rely on the crowding distance~\cite{crowding} to minimize multiple objectives under an accuracy constraint. The crowding distance is calculated for each solution in a Pareto front and is based on the distances between neighboring solutions in the objective space. The solutions with larger crowding distances are preferred in the selection process, as they represent areas of the objective space with lower solution density, and hence are more diverse and representative of the Pareto front.

During the search, we want to maximize the hardware efficiency of the adapted operator while keeping the difference between the loss of the original model $m$ and the model with the adapted operator, denoted as $m_{adapted}$, minimal. We add a small value, $\epsilon$, to ensure exploration. We fine-tune the network after adapting the operator for a few epochs. This fine-tuning is done with all the other operator's weights frozen. 

The operator's latency is computed with the difference between the original model's latency and the latency of the adapted model. The number of parameters can be reduced or increased by adding weight input to the computation graph. 

\begin{equation}
\begin{split}
    Min_{o} & (LAT(o), PARAM(o)) \\
    \text{subject to } & ACC(m_{adapted}) > ACC(m) - \epsilon
    \end{split}
    \label{eq:fitness}
\end{equation}

\section{Experiments}
\subsection{Experiment Settings and Implementation Details}\label{sec:exp_detail}
We first conducted our experiments on two edge devices: Raspberry Pi 3 Model B and Redmi Note 7S mobile phone. The Raspberry Pi 3 Model B is equipped with a Broadcom BCM2837 SoC with a 1.2 GHz quad-core ARM Cortex-A53 CPU, and 1GB RAM, and runs the Raspbian operating system. The Redmi Note 7S mobile phone is equipped with a Qualcomm Snapdragon 845 SoC with an octa-core CPU and 8GB RAM, running the Android 10 operating system.

To evaluate the performance of our proposed method, we used three popular deep learning models: ResNet18~\cite{resnet}, InceptionV3~\cite{inceptionv3}, and MobileNetV2~\cite{mobilenetv2}. We implemented our approach using Python 3.7 and the PyTorch 1.8.1 deep learning framework. All three architectures were initially trained for Imagenet. The experiment goal is to adapt them for edge devices by changing the most inefficient operators. We measured the accuracy of each model on the validation set and recorded each model's latency and energy consumption during inference. We averaged these numbers for 100 inferences to correctly estimate hardware efficiency. The latency and energy consumption are measured with an inference batch size of 1. {{For fine-tuning, we use SGD with a mini-batch size of 128. The learning rate is set to 0.003. We use a weight decay of 0.0001 and a momentum of 0.9.}}

The search is set to do 50 iterations per operator replacement. The stopping criterion is the modification of at least 10 layers in the model. The probability of mutation is set to 0.8 and the cross-over probability to 0.6. We use an epsilon of 1\%, i.e., assume that a 1\% drop in accuracy is acceptable. The epsilon should be tailored to the dataset and task at hand. Empirical tuning was done to select these values.

Due to the on-search fine-tuning and hardware efficiency computation on-device, our search takes about 1h04min. This time is highly practical as this adaptation is only done once. 

{{
\paragraph{Search Setup} 
The search is achieved on a much more compute-intensive setup. Our search was conducted using an NVIDIA GPU 3070, a high-performance graphics processing unit known for its advanced parallel computing capabilities. The GPU was connected to a powerful workstation equipped with an Intel Core i9 processor and 32 gigabytes of RAM, ensuring sufficient computational resources for the search process.
}}
\subsection{Optimizing an architecture for Edge Devices}
Table~\ref{tab:results} presents the overall hardware efficiency improvement achieved by applying GOS to the evaluated models on both edge devices. Our operator replacement method consistently outperformed the original models with an average speedup of 3.17. Notably, our search was able to find a variant that improved the accuracy of ResNet models by 6.13\% and 5.34\% for Raspberry Pi and Redmi Note 7S, respectively.

Interestingly, InceptionV3 was found to be unsuitable for deployment on Raspberry Pi due to its large network size. To tackle this issue, our search began optimizing by selecting operators that use the largest amount of parameters, which led to a reduction in the number of parameters and enabled the discovery of a deployable variant.

Although our search space does not directly optimize energy consumption, we observed that our models presented lower energy consumption due to the reduction in the number of parameters and operations.

Furthermore, GOS was able to find a variant of MobileNetV3 that is 2.2x faster with only a minor accuracy drop of 0.4\%, even though the original model was already optimized for mobile devices. Overall, our search consistently outperformed the original models, indicating the effectiveness of GOS in achieving hardware efficiency improvements.

{{
In comparing our strategy to other HW-NAS approaches, namely Once-for-All~\cite{ofa} and FBNetV3~\cite{fbnetv3}, we found that our operator replacement method yielded superior results. 
Our approach consistently outperformed Once-for-All and FBNetV3, showcasing an average speedup of 1.26. This performance advantage highlights the effectiveness of our method in optimizing neural architectures specifically for hardware constraints and further solidifies the value of GOS in achieving superior hardware efficiency improvements.
Note that our method can also be used as a specialization phase after the use of these high-level NAS. 
}}
\begin{table*}[h!]
\centering
\caption{Performance comparison of original models and adapted models on Raspberry Pi 3 and Redmi Note 7S}\label{tab:results}
\begin{tabular}{|c|l|l|c|p{1.5cm}|c|c|c|}
\hline
\begin{tabular}[c]{@{}l@{}}
Edge \\ Device
\end{tabular}                    & \textbf{Model}                        & \textbf{Variant} & \multicolumn{1}{l|}{\textbf{\# Parameters}} & \multicolumn{1}{p{1.5cm}|}{\textbf{Top-1 Accuracy (\%)}} & 
\multicolumn{1}{l|}{\textbf{Latency (ms)}} & \multicolumn{1}{l|}{\textbf{Energy (J)}} & \multicolumn{1}{l|}{\textbf{Speedup}} \\ \hline\hline
\multirow{6}{*}{\begin{turn}{90}\textbf{Raspberry Pi}\end{turn}}  & \multirow{2}{*}{\textbf{Resnet18~\cite{resnet}}}    & Original         & 11M                                         & 69.3                                              & 382.54                                     & 1320                                     & \multirow{2}{*}{5.76}          \\ \cline{3-7}
                                        &                                       & GOS              & 9.3M                                        & 75.43                                             & 66.32                                      & 220                                      &                                       \\ \cline{2-8} 
                                        & \multirow{2}{*}{\textbf{Inceptionv3~\cite{DBLP:conf/cvpr/SzegedyVISW16}}} & Original         & 25M                                         & 78.2                                              & -                                          & -                                        & \multirow{2}{*}{-}                    \\ \cline{3-7}
                                        &                                       & GOS              & 7.2M                                        & 79.47                                             & 101.3                                      & 438.3                                    &                                       \\ \cline{2-8} 
                                        & \multirow{2}{*}{\textbf{MobileNetV3~\cite{DBLP:conf/iccv/HowardPALSCWCTC19}}} & Original         & 2.9M                                        & 75.2                                              & 94.32                                      & 348                                      & \multirow{2}{*}{2.82}          \\ \cline{3-7}
                                        &                                       & GOS              & 2.9M                                        & 74.32                                             & 33.44                                      & 253                                      &                                       \\\cline{2-8} 
                                        &\multirow{2}{*}{\textbf{FBNetV3~\cite{fbnetv3}}} & Original         & 5.3M                                        & 79.1                                              & 25.4                                     & 238                                      & \multirow{2}{*}{1.52}          \\ \cline{3-7}
                                        &                                       & GOS              & 5.1M                                        & 83.4                                             & 16.7                                      & 187                                      &                                \\ 
                                        \cline{2-8} 
                                        &\multirow{2}{*}{\textbf{OFA~\cite{ofa}}} & Original         & 4.9M                                        & 74.2                                              & 22.3                                      & 211                                      & \multirow{2}{*}{1.16}          \\ \cline{3-7}
                                        &                                       & GOS              & 3.2M                                        & 79.3                                             & 19.2                                      & 204                                      &                                \\ 
                                        
                                        \hline\hline
\multirow{6}{*}{\begin{turn}{90}\textbf{Redmi Note 7S}\end{turn}} & \multirow{2}{*}{\textbf{Resnet18~\cite{resnet}}}    & Original         & 11M                                         & 69.3                                              & 93.43                                      & 119.4                                    & \multirow{2}{*}{4.51}           \\ \cline{3-7}
                                        &                                       & GOS              & 11.5M                                       & 74.64                                             & 20.7                                       & 78.8                                     &                                       \\ \cline{2-8} 
                                        & \multirow{2}{*}{\textbf{Inceptionv3~\cite{DBLP:conf/cvpr/SzegedyVISW16}}} & Original         & 25M                                         & 78.2                                              & 83.5                                       & 132.6                                    & \multirow{2}{*}{3.72}          \\ \cline{3-7}
                                        &                                       & GOS              & 23.4M                                       & 77.9                                              & 22.4                                       & 104.5                                    &                                       \\ \cline{2-8} 
                                        & \multirow{2}{*}{\textbf{MobileNetV3~\cite{DBLP:conf/iccv/HowardPALSCWCTC19}}} & Original         & 2.9M                                        & 75.2                                              & 76.3                                       & 76.54                                    & \multirow{2}{*}{2.23}          \\ \cline{3-7}
                                        &                                       & GOS              & 2.6M                                        & 74.8                                              & 34.2                                       & 78.43                                    &                                       \\ \cline{2-8} 
                                        &\multirow{2}{*}{\textbf{FBNetV3~\cite{fbnetv3}}} & Original         & 5.3M                                        & 79.1                                              & 21.6                                     & 67.9                                      & \multirow{2}{*}{1.18}          \\ \cline{3-7}
                                        &                                       & GOS              & 4.8M                                        & 81.4                                             & 18.3                                      & 87.3                                      &                                \\  \cline{2-8} 
                                        & \multirow{2}{*}{\textbf{OFA~\cite{ofa}}} & Original         & 4.6M                                        & 76.5                                              & 34.6                                       & 56.42                                    & \multirow{2}{*}{1.21}          \\ \cline{3-7}
                                        &                                       & GOS              & 3.7M                                        & 83.4                                             & 28.5                                       & 58.3                                    &                                       \\
                                        \hline
\end{tabular}
\end{table*}

\paragraph{Analysis of Resulting operations}
In this paragraph, we discuss the novel operators that were generated through our operator search method and the improvements they bring to the models. Table~\ref{tab:efficient_operators} presents the novel equations for the most efficient operators that replaced the standard convolution 2D, batch normalization, and activation functions. Table~\ref{tab:notation} summarizes the notations. Our discussion is focused on each device separately. 

In general, the last convolution 2D operators of the models are the most inefficient ones. Therefore, in all the models, we {{automatically optimized}} these operators {{using GOS}}. For the Raspberry Pi device, we modified these operators by adding a dilation rate to the convolutions, similar to dilated convolutions~\cite{dilated}. However, in our operator, the dilation rate is applied within the filter matrix itself, by adding 1, 2, or 3 zeroed columns between different columns of the filter matrix. This modification enables the operator to have a larger receptive field without increasing its size, which can be helpful in capturing features at different scales in an image. This operator is particularly efficient in Raspberry Pi, which has limited computational resources, as it reduces the number of operations needed to process the input.

On the other hand, for the Redmi Note 7S, the model's last convolution 2D operators were modified to a depthwise convolution~\cite{depthwise}. Similarly to Raspberry Pi, the dilated rate is applied here as well. The use of dilated filters and depthwise convolution allowed for an increase in hardware efficiency. It is worth noting that we did not start with a depthwise separable convolution, except for MobileNetV3. Instead, our operator search method converges to similar operations. In addition, for resnet18, the search applied a pooling layer at the end of the convolutions. This is done to further reduce the feature map size and enhance the latency. Interestingly, this did not impact negatively the accuracy. However a similar operator was tested on InceptionV3 and MobileNetV3 and a 5\%, 6.7\% drop in accuracy was seen. 

The first convolutions are particularly different from the last ones because of the input shape. In the first convolutions, the channel size is smaller, while the height and width of the feature maps are large. The opposite happens at the end of the model. The first convolutions, even in MobileNetV3, were turned into standard convolutions. The search only changed the hyperparameters of these operators, using 5x5 kernels for some and adding different padding. 

The generated batch normalization operation uses a polynomial regression to regress the batch normalization values after the standardization. By incorporating the polynomial regression into the batch normalization equation, this method can improve the accuracy of the normalization while maintaining a fast computation time.

The search algorithm almost never changes the activation functions, as they are usually fast and already efficient. However, we forced the model to change the activation and look for a more efficient version. The resulting equation is shown in table~\ref{tab:efficient_operators}. The equation is a leaky version of ReLU. When removing the activation functions from the list of instructions, the search failed to find a differentiable equation. 

\begin{table*}[h]
\centering
\caption{Efficient Operators Equations for Raspberry Pi and Redmi Note 7S}
\label{tab:efficient_operators}
\begin{tabular}{cccc}
\hline
\textbf{Device} & \textbf{Convolution2D} & \textbf{Batch Normalization} & \textbf{Activation} \\ \hline
Raspberry Pi & $\mathbf{y} = \sum_{i,j}\mathbf{w}{i,j}x{i+d_{i,j},j+d_{i,j}}$ & $\frac{x - \mu}{\sigma + \epsilon} \times \alpha(x - \mu)^3 + \beta(x - \mu) + \gamma$ & $max(0.01x,x)$ \\
Redmi Note 7S &  $\sum_{j=1}^{C_{in}}\sum_{r=1}^{H_{k}}\sum_{c=1}^{W_{k}}W_{i,j,r,c}\cdot I_{i+(r-1)d^{r}{j},j+(c-1)d^{c}{j},k}$& $\frac{x - \mu}{\sigma + \epsilon} \times \alpha(x - \mu)^3 + \beta(x - \mu) + \gamma$ & $max(0.03x,x)$ \\ \hline
\end{tabular}
\end{table*}

\begin{table}[h]
\centering
\caption{Notation Summary}
\label{tab:notation}
\begin{tabular}{cp{6.5cm}}
\hline
\textbf{Symbol} & \textbf{Description} \\ \hline
$C_{in}$ & Number of input channels \\ \hline
$C_{out}$ & Number of output channels \\ \hline
$H_{k}$ & Height of the kernel \\ \hline
$W_{k}$ & Width of the kernel \\ \hline
$K_{w}$ & Number of weights in the kernel \\ \hline
$d_{j}$ & Dilation rate for input channel $j$ \\ \hline
$d^{r}{j}$ & Dilation rate for input channel $j$ in the row direction \\ \hline
$d^{c}{j}$ & Dilation rate for input channel $j$ in the column direction \\ \hline
$I$ & Input tensor \\ \hline
$W$ & Weight tensor \\ \hline
$\gamma$ & Scaling parameter in batch normalization \\ \hline
$\beta$ & Bias parameter in batch normalization \\ \hline
$a_0$, $a_1$, $a_2$ & Polynomial coefficients in fast batch normalization \\ \hline
$x$ & Input to fast batch normalization \\ \hline
$\sigma$ & Standard deviation in batch normalization \\ \hline
$\epsilon$ & Small constant for numerical stability \\ \hline
\end{tabular}
\end{table}

\paragraph{Effect of number of instructions}
In the previous experiments, we fixed the maximum number of instructions per operator to 20. Here, we justify this value and analyze the effect of changing the number of instructions on operator generation, using the same search space and fitness evaluation as described in Section~\ref{sec:exp_detail}.

\begin{figure}
    \centering
    \includegraphics[width=0.5\textwidth]{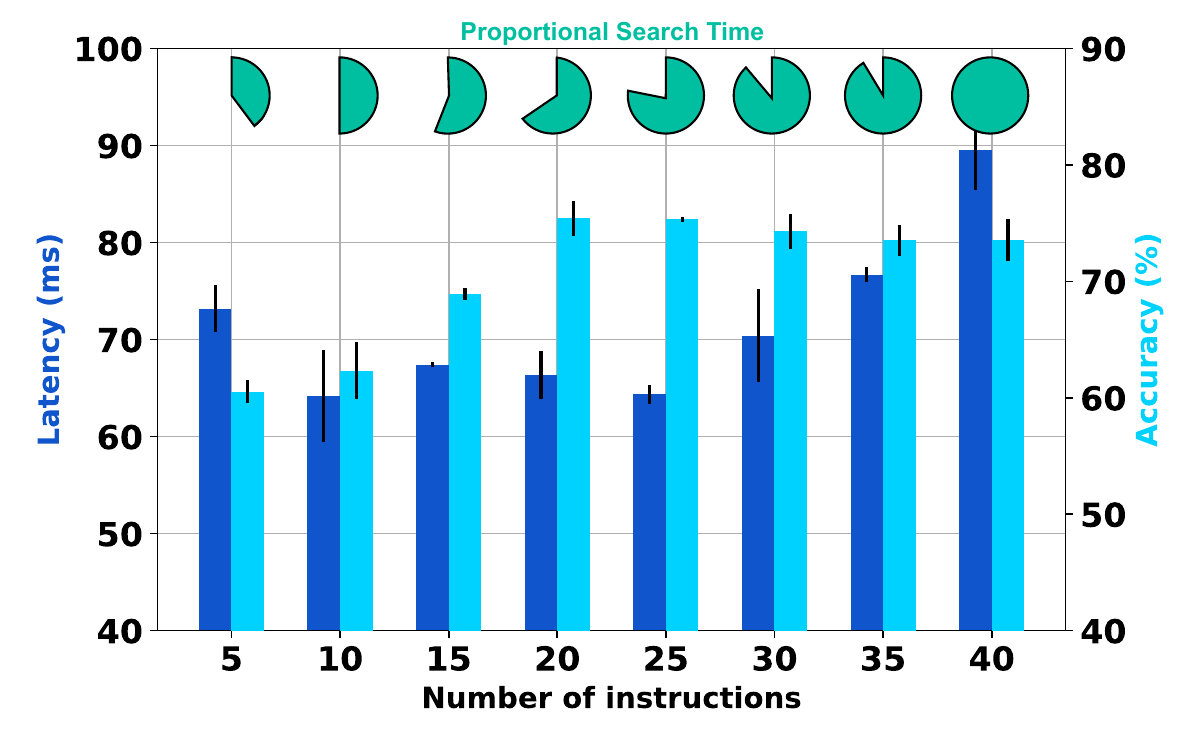}
    \caption{Tuning of the maximum number of instructions per operator while searching for resnet18 GOS variant on Raspberry Pi. }
    \label{fig:instruction}
\end{figure}

We varied the number of instructions used to define each operator ranging from 5 to 40 instructions with a step of 5 and compared the resulting architectures' performance. Specifically, we evaluated the accuracy and inference time of the architectures on the Imagenet dataset using the same hardware setup as in the previous experiments. Figure~\ref{fig:instruction} shows the results. 

The results showed that increasing the number of instructions used to define each operator generally leads to an improvement in the architectures' performance. This improvement stabilizes after 20 in which we obtain the results shown in table~\ref{tab:results}. Below 20, the operators become badly implemented and the accuracy drops. Above 25, the instruction set is too large and the operators apply redundant instructions which increases the latency. In addition, the search time highlighted in green increases with the increase of the maximum number of instructions per operator. This is due to the increased latency and fine-tuning time induced by more complex and redundant operators.

\subsection{Use Case: Pulse Rate Estimation}
The ability to estimate pulse rate continuously is a critical feature in heart attack detection. Estimating pulse rate is essential for measuring workout intensity during exercise and resting heart rate, which is often used to determine cardiovascular fitness. Using mobile wearable devices provides valuable insights into a wearer's health. Due to the limited hardware resources, the model needs to be small and fast to provide real-time results. This task also requires efficient processing of sensor data, which is a critical aspect of hardware-aware NAS. 
In addition sensed information is highly personal and requires edge inference with a fast and lightweight machine learning model. Typically, wearable devices sense an underlying signal, such as Photoplethysmography (PPG) and raw motion data to estimate pulse rate. Complex algorithms can then process raw data into various activity classifications or step counts. The algorithms used for this purpose range from simple linear regression to complex deep learning models, such as Convolutional Neural Networks (CNNs) and Long Short-Term Memory (LSTM) models.  In this use case, we focus on estimating the beats per minute (BPM) based on PPG and accelerometer raw data. 

During this experiment, the same previously used search hyperparameters are applied. 

\paragraph{Dataset}
For this task, we are using the Troika dataset~\cite{troika}. Troika is a publicly available dataset and contains measurements from three sensors to estimate the heart rate of the wearer: an ECG sensor, a PPG sensor, and an accelerometer sensor. The dataset was collected in a study where participants were asked to perform a set of activities while wearing the sensors, including running, cycling, and sitting. The dataset contains 12 recordings from 8 participants, aged 18 to 35, with each recording lasting 5 minutes. The ground truth heart rate for each recording was obtained from the ECG sensor, which is considered the most accurate method for measuring heart rate. 

 \paragraph{Wrist-band device}
Our latency results were extracted from a Xiaomi Mi Smart Band 6. The wristband is composed of a built-in PPG biosensor and a 3-axis accelerometer sensor which extracts the input to our algorithm. The operating system installed is Android 6.0. And the battery of the system was at maximum when extracting the latencies. 
 
\begin{table*}[ht!]
\centering
\caption{Results of Average Absolute Error for Pulse Rate estimation on TROIKA Dataset~\cite{troika}}
\label{tab:pulse_rate}
\begin{tabular}{|ll|p{1cm}|p{1cm}|c|c|c|c|}
\hline
\multicolumn{1}{|l|}{\textbf{Action}}                      & \textbf{Subject}                      & \multicolumn{1}{l|}{\textbf{\begin{tabular}[c]{@{}l@{}}RF\\Optimized \\ (Ours)\end{tabular}}} & \multicolumn{1}{l|}{\textbf{\begin{tabular}[c]{@{}l@{}}PPG\_NAS\\Optimized \\ (Ours)\end{tabular}}} & \multicolumn{1}{l|}{\textbf{RF\_Model}} & \multicolumn{1}{l|}{\textbf{\begin{tabular}[c]{@{}l@{}}PPG\_NAS\\\cite{ppgnas}\end{tabular}}} & \multicolumn{1}{l|}{\textbf{\begin{tabular}[c]{@{}l@{}}CNN-LSTM\\\cite{CNN-LSTM}\end{tabular}}} & \multicolumn{1}{l|}{\textbf{\begin{tabular}[c]{@{}l@{}}DeepHeart\\\cite{deepheart}\end{tabular}}} \\ \hline
\multicolumn{1}{|l|}{T1}                                   & 1                                     & 1.43                                                                                          & 0.8                                                                                                 & 5.9                                     & 0.95                                   & 0.47                                   & 1.47                                    \\ \hline
\multicolumn{1}{|l|}{T1}                                   & 2                                     & 2.08                                                                                          & 1.33                                                                                                & 1.57                                    & 1.22                                   & 3.88                                   & 2.94                                    \\ \hline
\multicolumn{1}{|l|}{T1}                                   & 3                                     & 3.76                                                                                          & 0.06                                                                                                & 4.24                                    & 0.43                                   & 1.52                                   & 0.47                                    \\ \hline
\multicolumn{1}{|l|}{T1}                                   & 4                                     & 2.08                                                                                          & 0.69                                                                                                & 8.68                                    & 0.69                                   & 2.31                                   & 1.02                                    \\ \hline
\multicolumn{1}{|l|}{T1}                                   & 5                                     & 1.05                                                                                          & 0.83                                                                                                & 2.74                                    & 0.72                                   & 1.72                                   & 2.66                                    \\ \hline
\multicolumn{1}{|l|}{T1}                                   & 6                                     & 4.24                                                                                          & 0.72                                                                                                & 4.49                                    & 0.49                                   & 1.47                                   & 0.75                                    \\ \hline
\multicolumn{1}{|l|}{T1}                                   & 7                                     & 1.51                                                                                          & 0.71                                                                                                & 4.8                                     & 0.99                                   & 2.85                                   & 3.45                                    \\ \hline
\multicolumn{1}{|l|}{T1}                                   & 8                                     & 2.57                                                                                          & 1.3                                                                                                 & 10.81                                   & 0.87                                   & 2.18                                   & 2.48                                    \\ \hline
\multicolumn{1}{|l|}{T1}                                   & 9                                     & 3.87                                                                                          & 1.44                                                                                                & 7.41                                    & 1.06                                   & 4.9                                    & 0.54                                    \\ \hline
\multicolumn{1}{|l|}{T1}                                   & 10                                    & 4.49                                                                                          & 0.98                                                                                                & 11.18                                   & 0.64                                   & 0.34                                   & 0.72                                    \\ \hline
\multicolumn{1}{|l|}{T1}                                   & 11                                    & 3.7                                                                                           & 0.87                                                                                                & 20.16                                   & 1.01                                   & 4.46                                   & 1.06                                    \\ \hline
\multicolumn{1}{|l|}{T1}                                   & 12                                    & 2.63                                                                                          & 0.77                                                                                                & 5.37                                    & 0.67                                   & 1.79                                   & 0.73                                    \\ \hline
\multicolumn{1}{|l|}{T2}                                   & 13                                    & 5.24                                                                                          & 1.96                                                                                                & 5.56                                    & 1.62                                   & 3.01                                   & 4.8                                     \\ \hline
\multicolumn{1}{|l|}{T2}                                   & 14                                    & 4.2                                                                                           & 1.84                                                                                                & 23.32                                   & 1.95                                   & 7.6                                    & 2.94                                    \\ \hline
\multicolumn{1}{|l|}{T2}                                   & 15                                    & 9.89                                                                                          & 1.24                                                                                                & 9.92                                    & 0.59                                   & 1.58                                   & 0.11                                    \\ \hline
\multicolumn{1}{|l|}{T3}                                   & 16                                    & 5.22                                                                                          & 0.57                                                                                                & 5.49                                    & 0.61                                   & 0.9                                    & 1.63                                    \\ \hline
\multicolumn{1}{|l|}{T3}                                   & 17                                    & 1.32                                                                                          & 1.14                                                                                                & 1.58                                    & 1.32                                   & 6.1                                    & 1.84                                    \\ \hline
\multicolumn{1}{|l|}{T3}                                   & 18                                    & 1.59                                                                                          & 0.48                                                                                                & 5.98                                    & 0.55                                   & 0.31                                   & 1.64                                    \\ \hline
\multicolumn{1}{|l|}{T3}                                   & 19                                    & 0.2                                                                                           & 0.54                                                                                                & 0.61                                    & 0.47                                   & 0.12                                   & 0.18                                    \\ \hline
\multicolumn{1}{|l|}{T3}                                   & 21                                    & 2.52                                                                                          & 1.18                                                                                                & 4.65                                    & 0.39                                   & 0.38                                   & 0.06                                    \\ \hline
\multicolumn{1}{|l|}{T3}                                   & 22                                    & 0.83                                                                                          & 0.93                                                                                                & 4.23                                    & 0.83                                   & 1.26                                   & 2.25                                    \\ \hline
\multicolumn{1}{|l|}{T2}                                   & 23                                    & 2.88                                                                                          & 1.54                                                                                                & 8.17                                    & 1.38                                   & 4.26                                   & 0.94                                    \\ \hline
\multicolumn{2}{|l|}{\textbf{All}}                                                                 & 3.13                                                                                          & 1.03                                                                                                & 7.34                                    & 0.93                                   & 2.51                                   & 1.68                                    \\ \hline
\multicolumn{2}{|l|}{\textbf{Latency (ms)}}                                                        & 2.38                                                                                          & 2.68                                                                                                & 1.64                                    & 5.6                                    & 11.8                                   & 13.54                                   \\ \hline
\multicolumn{2}{|l|}{\textbf{\begin{tabular}[c]{@{}l@{}}Number of \\ parameters (M)\end{tabular}}} & 0.08                                                                                          & 0.564                                                                                               & 0.02                                    & 1.1                                    & 3.3                                    & 4.4                                     \\ \hline
\end{tabular}
\end{table*}
\paragraph{Preprocessing}
In models designed for pulse rate estimation, preprocessing of the raw data plays a critical role. Preprocessing is used to clean and enhance raw photoplethysmography (PPG) signals before the actual analysis. The dataset was divided into subjects, and each subject included the PPG and accelerometer as time-series data. The PPG and individual accelerometer signals undergo bandpass filtering with a range of 40BPM to 240BPM (0.66Hz and 40Hz respectively) to eliminate irrelevant signals. In the next step, characteristic frequencies are calculated for each signal within 8-second windows, with a 6-second overlap between adjacent windows. The characteristic frequencies for the PPG data are then compared against the dominant characteristic frequencies for each accelerometer axis per time window. It is possible that a dominant PPG frequency may be similar to accelerometer frequencies. To ensure an accurate estimation of pulse rate, we compare up to 10 PPG frequencies that potentially represent the pulse rate to find one that is not similar to accelerometer frequencies. In cases where no alternate frequency is found, we select the PPG frequency with the largest magnitude as the pulse rate. 

\paragraph{Models}
We optimize two models using our methodology: 
\begin{itemize}
    \item \textit{RF\_Model}: We first optimize a simple machine-learning model consisting of two blocks. The first block consists of a bandpass filter and a Fourier transform.  The PPG signal contains information about the blood flow in the capillaries. This signal is a combination of various frequencies, including the pulse rate. By applying a bandpass filter to the PPG signal, frequencies outside the range of interest are eliminated. The Fourier transform is then applied to the filtered signal to extract the characteristic frequencies that correspond to the pulse rate. This process helps to remove noise and artifacts from the signal and facilitates accurate estimation of the pulse rate. The second block consists of a random forest regressor~\cite{randomforest}. While this first model is fast, it is not optimal in terms of performance. 
    \item \textit{PPG\_NAS\_Model}: PPG\_NAS~\cite{ppgnas} is a dedicated NAS for PPG signal analysis and pulse rate estimation. The authors generated an optimized model for pulse rate estimation. The model consists of a 1D convolution layer followed by 2 LSTM layers and a final fully-connected layer. The architecture is designed to minimize the number of parameters and maximize the accuracy of pulse rate estimation. This is a state-of-the-art model in terms of the accuracy of the regression and hardware efficiency on wristband devices. 
\end{itemize}

\paragraph{Results}
Table~\ref{tab:pulse_rate} shows the overall average absolute error (AAE) results on multiple subjects and under different actions: running ($T1$), cycling ($T2$), and sitting ($T3$), as well as their latency and number of parameters. We additionally compare our results to other state-of-the-art models, namely PPG\_NAS~\cite{ppgnas}, CNN-LSTM~\cite{CNN-LSTM}, and DeepHeart~\cite{deepheart}. 

Over all subjects and actions, our models outperform their state-of-the-art counterparts with a lesser number of parameters and faster latency. 

\begin{figure*}[h!]
    \centering
    \includegraphics[width=0.8\textwidth]{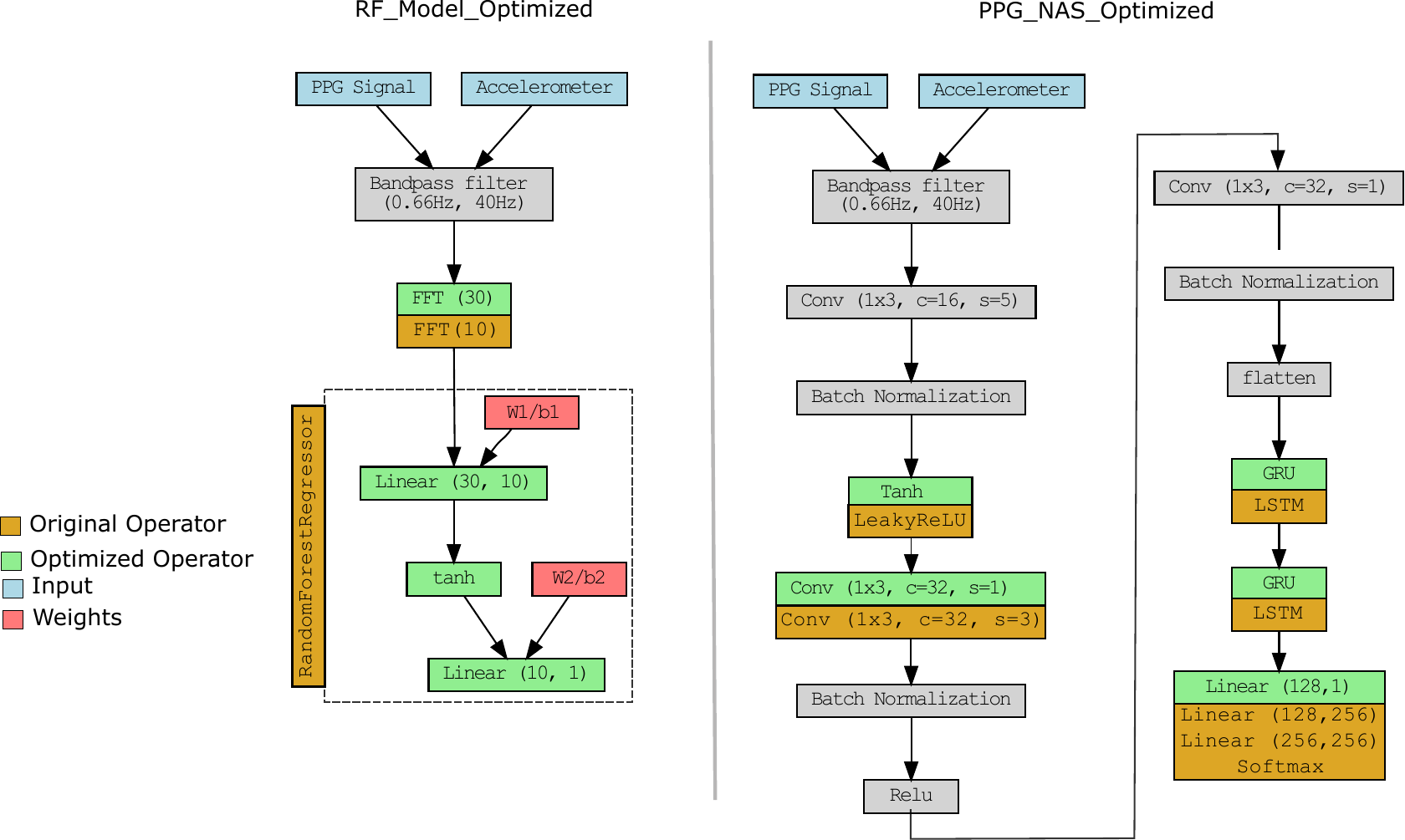}
    \caption{Pulse Rate Estimation final Models. We do not display the weights node for PPG\_NAS for the sake of clarity.}
    \label{fig:pulse_rate_models}
\end{figure*}

Figure~\ref{fig:pulse_rate_models} shows the final architectures proposed by our Grassroots operator search. The optimized final models of \\ RF\_Model\_optimized and PPG\_NAS\_optimized were obtained by modifying their respective base models. RF\_Model\_optimized underwent two stages of modifications. First, the fast Fourier transform was altered to extract 30 points instead of the 10 peaks in the original model. Additionally, a linear layer was added to act as a smoother filter that selects the 10 most significant peaks. In the second stage, the random forest regressor was replaced with a multi-layer perceptron (MLP) using a hyperbolic tangent (tanh) activation function. While the RandomForestRegressor is highly efficient, it reduces the model's accuracy, so the search favored MLP. As for PPG\_NAS\_optimized, the LSTM layers in the original model were replaced with gated recurrent units (GRUs), which use fewer parameters and do not affect the performance. The convolution layer was also modified to use a dilated-like convolution (as shown in the table), and the final linear function was adjusted to have no activation at the end. These optimizations resulted in more accurate and efficient models for pulse rate estimation.

\section{Conclusion}
\label{conclusion}
In conclusion, we have presented a novel approach for optimizing neural network architectures for resource-constrained devices. Our approach leverages the use of mathematical equations to replace common operations such as convolution, batch normalization, and activation functions with more efficient ones, resulting in models that are optimized for low-power devices such as Raspberry Pi and mobile phones. We demonstrated the effectiveness of our approach through experiments on popular architectures, including ResNet18, InceptionV3, and MobileNetV3, achieving significant improvements in inference time and energy consumption compared to the original models. Additionally, we applied GOS to a real-world healthcare problem, namely pulse rate estimation, in which we presented a 2x faster network with a 0.12 average error drop. Overall, our results highlight the potential of our approach for creating efficient neural networks for resource-constrained devices.



\newpage
\bibliographystyle{elsarticle-num-names} 
\bibliography{references}






\end{document}